# Approximate Equalities on Rough Intuitionistic Fuzzy Sets and an Analysis of Approximate Equalities

B. K. Tripathy[1], G. K. Panda[2]

[1] SCSE, VIT University
Vellore, Tamilnadu 632014, India
*tripathybk@rediffmail.com*

[2] Department of CSE and IT, M.I.T.S.,
Rayagada, Odisha 765017, India
*gkpmail@sify.com*

**Abstract**

In order to involve user knowledge in determining equality of sets, which may not be equal in the mathematical sense, three types of approximate (rough) equalities were introduced by Novotny and Pawlak ([8, 9, 10]). These notions were generalized by Tripathy, Mitra and Ojha ([13]), who introduced the concepts of approximate (rough) equivalences of sets. Rough equivalences capture equality of sets at a higher level than rough equalities. More properties of these concepts were established in [14]. Combining the conditions for the two types of approximate equalities, two more approximate equalities were introduced by Tripathy [12] and a comparative analysis of their relative efficiency was provided. In [15], the four types of approximate equalities were extended by considering rough fuzzy sets instead of only rough sets. In fact the concepts of leveled approximate equalities were introduced and properties were studied. In this paper we proceed further by introducing and studying the approximate equalities based on rough intuitionistic fuzzy sets instead of rough fuzzy sets. That is we introduce the concepts of approximate (rough) equalities of intuitionistic fuzzy sets and study their properties. We provide some real life examples to show the applications of rough equalities of fuzzy sets and rough equalities of intuitionistic fuzzy sets.

***Keywords:*** *Approximate equality, Leveled approximate equality, Fuzzy set, Rough set, Rough Fuzzy set, Intuitionistic Fuzzy Sets, Rough Intuitionistic Fuzzy Sets, Accuracy measures*

## 1. Introduction

The notions of fuzzy sets and rough sets were introduced by Zadeh [16] and Pawlak [3] respectively in order to model imperfect knowledge. In the beginning they were supposed to be competing models. But later on it was established that actually they are complementary to each other and the hybrid models taking both these models together are still more effective for applications than the individual ones [2]. Intuitionistic fuzzy sets are better models than fuzzy sets and are more realistic in a sense. So hybrid models of rough sets with intuitionistic fuzzy sets are better models than the rough fuzzy hybrid models.

The equality of sets in the mathematical sense was found to be very restrictive in real life applications and so three notions of approximate equalities were introduced by Novotny and Pawlak ([8, 9, 10]) which depend upon the available knowledge about the structure of the universe in which they are defined. These notions were further extended to define the notions of rough equivalences and some real life examples are provided in [13, 14] to illustrate the superiority of these concepts. Tripathy [12] introduced two more types of approximate equalities of sets and an analysis of the four kinds of approximate equalities was made regarding their applicability and efficiency.

The four types of approximate equalities of sets introduced in [12] were recently extended to the context of fuzzy sets by Tripathy et al [15] by using the hybrid notion of rough fuzzy sets. In fact, they have introduced the notions of leveled approximate equalities. So that depending upon the requirements of the user and the application at hand the level of approximate equalities can be defined by the user and interpreted.

Several properties of approximate equalities established in [8, 9, 10] and the corresponding replacement properties were analysed in [13] and it was shown that some of the statements were incorrect. In [13] attempts were made to extend these properties to the generalized situation of rough equivalences. Using the rough equivalences validity of some basic algebraic properties involving union, intersection and complementation of rough sets were established in [14].

The structure of this paper is as follows. In section 2, we introduce some definitions and notations, which are to be used in this paper. In section 3, the basic approximate equalities of crisp sets are discussed. In section 4, the approximate equalities of fuzzy sets are presented. Here, we modify the definition given in [15] so that it is compatible with the definition of rough fuzzy sets provided by Dubois and Prade [2]. Also, we provide some examples to illustrate the applicability of the notions in real life situations. In section 5 we introduce the approximate equalities of intuitionistic fuzzy sets using the notion of rough intuitionistic fuzzy sets. Here again, we provide examples from real life to illustrate the application of the concepts introduced. Many of the properties of the new notions which extend the corresponding properties of the earlier case are established. In section 6, some concluding remarks are presented and section 7 is a compilation of bibliographic references referred during the compilation of this article.

## 2. Definitions and Notations

In this section we introduce some basic concepts which are to be used in this paper. We start with the definition of basic rough sets as introduced by Pawlak [3].

### 2.1 Rough Sets

Let $U$ be a universe of discourse and $R$ be an equivalence relation over $U$. By $U/R$ we denote the family of all equivalence classes of $R$, referred to as categories or concepts of $R$ and the equivalence class of an element $x \in U$, is denoted by $[x]_R$. By a knowledge base, we understand a relational system $K = (U, R)$, where $U$ is as above and $R$ is a family of equivalence relations over $U$. For any subset $P(\neq \phi) \subseteq R$, the intersection of all equivalence relations in $P$ is denoted by IND (P) and is called the indiscernibility relation over $P$. Given any $X \subseteq U$ and $R \in IND(K)$, we associate two subsets,
$\underline{R}X = \bigcup\{Y \in U/R : Y \subseteq X\}$ and
$\overline{R}X = \bigcup\{Y \in U/R : Y \cap X \neq \phi\}$, called the R-lower and R-upper approximations of X respectively.
The R-boundary of $X$ is denoted by $BN_R(X)$ and is given by $BN_R(X) = \overline{R}X - \underline{R}X$. We say that $X$ is rough with respect to $R$ if and only if $\underline{R}X \neq \overline{R}X$, equivalently $BN_R(X) \neq \phi$. $X$ is said to be R-definable if and only if $\underline{R}X = \overline{R}X$ or $BN_R(X) = \phi$.

### 2.2 Fuzzy Sets

As mentioned in the introduction, fuzzy set introduced by Zadeh [16] is one of the early approaches to capture vagueness in concepts. In the fuzzy set approach every member x of a set $X \subseteq U$ is associated with a grade of membership, which we denote by X(x) and is a real number lying in [0, 1]. The set of all functions from U to the unit interval [0, 1], is called the fuzzy power set of U and is denoted by F(U). It follows that P(U) $\subseteq$ F(U).

### 2.3 Rough Fuzzy Sets

In the beginning when rough sets were introduced by Pawlak in the early 1980s, it was supposed to be a rival to the theory of fuzzy sets. But it was established by Dubois and Prade [2] that instead of being rival theories, they complement each other. In fact they combined these two models to develop the hybrid models of fuzzy rough sets and rough fuzzy sets.
The notion of rough fuzzy sets was introduced by Dubois and Prade [2] as follows.
**Definition 2.3.1:** Let (U, R) be an approximation space and $U/R = \{X_1, X_2, ... X_n\}$. Then for any X $\in$ F(U), $\underline{R}X$ and $\overline{R}X$, the lower and upper approximations of X with respect to R are fuzzy sets in U/R. That is, $\underline{R}X, \overline{R}X : U/R \to [0,1]$, such that
(2.3.1)  $(\underline{R}X)(X_j) = \inf_{y \in X_j} X(y)$. and
(2.3.2)  $(\overline{R}X)(X_j) = \sup_{y \in X_j} X(y)$, for all $j = 1,2,...n$.
The pair $(\underline{R}X, \overline{R}X)$ is called a rough fuzzy set associated with X
**Definition 2.3.2:** Let (U, R) be an approximation space and $U/R = \{X_1, X_2, ... X_n\}$. Then for any X $\in$ F(U), we define two fuzzy sets $(\underline{R}X)_F$ and $(\overline{R}X)_F : U \to [0,1]$ such that for any $x \in X_i, i = 1,2,...n;$ we have
(2.3.3) $(\underline{R}X)_F(x) = (\underline{R}X)(X_i)$ and
(2.3.4) $(\overline{R}X)_F(x) = (\overline{R}X)(X_i)$.

### 2.4 Intuitionistic Fuzzy Sets

In the intuitionistic fuzzy set approach every member x of a set $X \subseteq U$ is associated with a grade of membership and a grade of nonmembership, which we denote by MX(x) and NX(x) respectively. For all x $\in$ U both MX(x) and NX(x) are real number lying in [0, 1], such that 0 $\leq$ MX(x) +NX(x) $\leq$ 1. The set of all functions from U to J, where J = {(m, n) | m, n $\in$ [0, 1] and 0 $\leq$ m+ n $\leq$ 1}, is called the intuitionistic fuzzy power set of U and is denoted by IF(U). It follows that P(U) $\subseteq$ F(U) $\subseteq$ IF(U).

### 2.5 Rough Intuitionistic Fuzzy Sets

Extending the notion of rough fuzzy sets introduced by Dubois and Prade, rough intuitionistic fuzzy sets can be defined as follows.
Let (U, R) be an approximation space and U/R $= \{X_1, X_2, ... X_n\}$. Then for any X $\in$ IF(U), $\underline{R}X$ and $\overline{R}X$, the lower and upper approximations of X with respect to R are intuitionistic fuzzy sets in U. That is,
$\underline{R}X, \overline{R}X : U/R \to [0,1]$, such that
(2.5.1)
$M(\underline{R}X)(X_j) = \inf_{y \in X_j} MX(y)$ and $N(\underline{R}X)(X_j) = \sup_{y \in X_j} NX(y)$ and
(2.5.2)
$M(\overline{R}X)(X_j) = \sup_{y \in X_j} MX(y)$ and $N(\overline{R}X)(X_j) = \inf_{y \in X_j} NX(y)$
For all j = 1,2, ..n.

We define
$\underline{R}X = (M(\underline{R}X), N(\underline{R}X))$ and $\overline{R}X = (M(\overline{R}X), N(\overline{R}X))$ .

The pair $(\underline{R}X, \overline{R}X)$ is called
a rough intuitionistic fuzzy set associated with X.

**Definition 2.5.2:** Let (U, R) be an approximation space and $U/R = \{X_1, X_2, ...X_n\}$. Then for any $X \in IF(U)$, we define two intuitionistic fuzzy sets $(\underline{R}X)_{IF}$ and $(\overline{R}X)_{IF}$ with their membership and non-membership functions defined for any $x \in X_i, i = 1, 2, ...n$; by
(2.5.3)
$M(\underline{R}X)_{IF}(x) = M(\underline{R}X)(X_i)$ and $N(\underline{R}X)_{IF}(x) = N(\underline{R}X)(X_i)$
(2.5.4)
$M(\overline{R}X)_{IF}(x) = M(\overline{R}X)(X_i)$ and $N(\overline{R}X)_{IF}(x) = N(\overline{R}X)(X_i)$.

## 3. Approximate equalities of Sets

In the introduction, we have mentioned the necessity of dealing with approximate equalities of sets. Basically these notions are dependent upon the knowledge of the assessor about the universe. As noted by Pawlak ([2], p.26 ), all these approximate equalities of sets are of relative character, that is things are equal or not equal from our point of view depending on what we know about them. So, in a sense the definition of rough equality refers to our knowledge about the universe. In fact, Novotny and Pawlak [8, 9, 10] introduced three types of approximate equalities called bottom rough equality, top rough equality and the rough equality. Extending these early notions of approximate equalities three other types were introduced in [14] and [12]. We summarise these four types of approximate equalities in Table 3.1 below.

| Upper approximation / Lower approximation | $\overline{R}X = \overline{R}Y$ | $\overline{R}X$ and $\overline{R}Y$ are $U$ or $\neq U$ together |
|---|---|---|
| $\underline{R}X = \underline{R}Y$ | Rough Equalities | Approximate rough equalities |
| $\underline{R}X$ and $\underline{R}Y$ are $\phi$ or $\neq \phi$ together | Approximate rough equivalences | Rough Equivalences |

Several properties of rough equalities were established by Novotny and Pawlak [8, 9, 10 and also see 4]. The properties obtained from these properties by interchanging bottom and top approximate equalities are called replacement properties [4, 11, 12, 13, 14]. It was mentioned in [4] that these replacement properties do not hold in general. But, it is established in [11] that some of these properties actually hold and the others hold under suitable conditions.

### 3.1 Comparisons of approximate equalities

In [12] the following comparative analysis of the four types of approximate equalities was made.

**3.1.1** The condition that two sets are lower approximately equal if and only if the two sets have the same lower approximation is satisfied in only rare and restricted cases. Since we are using this property in case of both rough equal and approximately rough equal definitions, these two cases of rough equalities seem to have lesser utility than the corresponding rough equivalences

**3.1.2** The condition that the two upper approximations be equal provides freedom to define equality in a very approximate sense and is quite general than these two being equal to U or not simultaneously. But, sometimes it seems to be unconvincing.

**3.1.3** The concept of approximate rough equivalence is neither unconvincing nor unnatural. This is the most natural and best among the four types.

**3.1.4** The fourth type of approximate rough equality happens to be the worst among the four types of approximate equalities considered.

## 4. Approximate equalities using rough fuzzy sets

It is known that two fuzzy sets X and Y are equal if and only if X(x) = Y(x) for all x in U. Like the approach through which we could incorporate human knowledge using rough sets to define four types of rough equalities for crisp sets, four types of rough equalities were defined for fuzzy sets in [15]. It is worth noting that, here we provide a modified version of the definitions here. This definition is more appropriate in the context of the definition of rough fuzzy set provided here. This does not affect the results established or the examples used for illustration in [17] but it makes it more rigorous and authentic. We present these notions below.

Let X and Y be any two fuzzy sets on U. We denote their lower approximations by $\underline{R}X$, $\underline{R}Y$ and upper approximations by $\overline{R}X$, $\overline{R}Y$ respectively. Let $(\underline{R}X)_F, (\underline{R}Y)_F, (\overline{R}X)_F$ and $(\overline{R}Y)_F$ be the corresponding fuzzy sets on U associated with these approximations. For any fuzzy set A, the $\alpha - cut$ of A is denoted by $A_\alpha$ and it consists of elements of U having membership value greater than $\alpha$. The $\alpha - cut$ of a fuzzy set is a crisp set and so the equalities in the table below are crisp set equalities, which represent the four types of rough fuzzy equalities. These equalities introduce $\alpha$-level approximate equalities for fuzzy sets. This provides a control with the user to consider equalities as per the requirement. When X and Y are crisp sets, the concepts reduce to their corresponding notions of rough approximate equalities as the $\alpha - cuts$ for any fixed $\alpha > 0$ contain those elements which are in the sets in their crisp form only. So that table 4.1 below reduces to table 3.1 above.

Table 4.1

| Upper Approximation → Lower Approximation ↓ | $((\overline{R}X)_F)_\alpha = ((\overline{R}Y)_F)_\alpha$ | $((\overline{R}X)_F)_\alpha$ and $((\overline{R}Y)_F)_\alpha$ are U or not U together |
|---|---|---|
| $((\underline{R}X)_F)_\alpha = ((\underline{R}Y)_F)_\alpha$ | $\alpha$ – Rough Fuzzy Equality | $\alpha$ – Approximate Rough Fuzzy Equality |
| $((\underline{R}X)_F)_\alpha$ and $((\underline{R}Y)_F)_\alpha$ are $\phi$ or not $\phi$ together | $\alpha$ – Approximate Rough Fuzzy Equivalence | $\alpha$ – Rough Fuzzy Equivalence |

In [15] some examples were provided in order to illustrate the relative efficiencies of the four types of approximate equalities of fuzzy sets.

Properties of the four types of approximate $\alpha$ – rough fuzzy equalities were obtained in [15]. These properties extend the properties of rough equalities established in [8, 9, 10] and rough equivalences established in [13].

4.1 Comparisons of rough fuzzy equalities

We provide below a comparison of rough equalities of fuzzy sets, which is parallel to that for rough equalities of crisp sets.

**4.1.1** The condition that two fuzzy sets are lower approximately equal if and only if the two sets have the same $\alpha$ – support set of lower approximation is satisfied in only those rare and restricted cases where the members have same membership values from and after $\alpha$. Since we are using this property in case of both rough fuzzy equal and approximately rough fuzzy equal definitions, these two cases of rough fuzzy equalities seem to have lesser utility than the corresponding rough fuzzy equivalences

**4.1.2** The condition that the two upper approximations be equal provides freedom to define equality in a very approximate sense and is quite general than these two being equal to U or not simultaneously. As illustrated in the above examples, the later restriction sometimes seems to be less unconvincing.

**4.1.3** The concept of approximate rough fuzzy equivalence is neither unconvincing nor unnatural. This is the most natural and best among the four types as provided through examples 3.1 and 3.2 above.

**4.1.4** The fourth type of approximate rough fuzzy equality happens to be the worst among the four types of approximate equalities considered.

**4.1.5 Real life Examples**
In this section we provide two real life examples to illustrate the rough fuzzy approximate equalities.

**Example 1:** Let us take the universe U as the set of people in a town. We define a relation R over U as two people x and y are R-related to each other if and only if they belong to the same ward. This relation is clearly an equivalence relation over U and obviously decomposes U into equivalence classes, which are people in individual wards.

Basing upon their financial status, we define some fuzzy sets over U as Very poor, Poor, Lower middle class, Middle class, Upper middle class, Rich and Very rich. These are fuzzy sets and can be defined through fuzzy membership functions with obvious overlapping.

Let us consider the uniformity of the distribution of any two strata of people from the above over the wards. Consider for example Rich and middle class. If the upper approximations of these two fuzzy sets with respect to R are same then they are equally distributed over all the wards containing them. On the other hand if both their upper approximations are U, then these two classes of people are equally distributed over all the wards in the town.

**Example 2:** Let us consider the marks of students in undergraduate courses in a state. We define a relation R over U two students are R-related to each other if and only if they belong to the same college. This relation is an equivalence relation over the set U and decomposes it into equivalence classes, which are students in individual colleges.

Basing upon their performance in the final examination we define some fuzzy sets over U as Very bad, Bad, Average, Good and Very good students in the state. These fuzzy sets can be defined over the interval of percentage of marks in the interval of [0, 100].

Here we would like to find the uniform distribution of a particular category of students over different colleges of the state. Suppose we want to find the distribution of Average and Good students. Then we take the upper approximation of these two sets. If these are equal and not equal to U then the distribution is equally uniform over these colleges. If both are equal and each is equal to U then the distribution is equally uniform over all the colleges in the state. If these are not equal then we can say that the distribution is not equally uniform. We can have similar analysis over any pair of these fuzzy sets.

# 5. Approximate equalities using rough intuitionistic fuzzy sets

Following the patterns for approximate equalities we can define four types of approximate equalities for intuitionistic fuzzy sets as follows.

Here, we take X and Y to be intuitionistic fuzzy sets. $\underline{R}X = (M\underline{R}X, N\underline{R}X)$ and $\underline{R}Y = (M\underline{R}Y, N\underline{R}Y)$ are the lower approximations of X and Y. $\overline{R}X = (M\overline{R}X, N\overline{R}X)$ and $\overline{R}Y = (M\overline{R}Y, N\overline{R}Y)$ are the upper approximations of X and Y respectively. Following definition 2.5.2, let the corresponding intuitionistic fuzzy sets on U be $(\underline{R}X)_{IF}, (\underline{R}Y)_{IF}, (\overline{R}X)_{IF}$ and $(\overline{R}Y)_{IF}$ respectively. For any intuitionistic fuzzy set A, we denote the $(\alpha,\beta)-cut$ of A by $A_{\alpha,\beta} = \{x : MA(x) > \alpha \text{ and } NA(x) < \beta\}$.

**Note 5.1** When A is a fuzzy set, we have NA(x) = 1- MA(x). So, taking $\beta = 1-\alpha$ we see that $(A)_{\alpha,\beta} = A_\alpha$. Hence, table 5.1 reduces to table 4.1. In general, we have two control parameters $\alpha$ and $\beta$ with the user. Since $(\alpha,\beta) \in J$, we have infinite number of ways to talk about the approximate equalities of two intuitionistic fuzzy sets under four major categories in table 5.1. Further, when $\alpha = 0$ we get the rough equalities for crisp sets. Since the $(\alpha,\beta)-cut$ of an intuitionistic fuzzy set is a crisp set, the equalities in the table below are crisp set equalities.

Table 5.1

| Upper Approximation → / Lower Approximation ↓ | $((\overline{R}X)_{IF})_{\alpha,\beta} = ((\overline{R}Y)_{IF})_{\alpha,\beta}$ | $(\overline{R}X)_{IF})_{\alpha,\beta}$ and $((\overline{R}Y)_{IF})_{\alpha,\beta}$ are U or not U together |
|---|---|---|
| $((\underline{R}X)_{IF})_{\alpha,\beta} = ((\underline{R}Y)_{IF})_{\alpha,\beta}$ | $(\alpha,\beta)$ – Rough Intuitionistic Fuzzy Equality | $(\alpha,\beta)$ – Approximate Rough Intuitionistic Fuzzy Equality |
| $((\underline{R}X)_{IF})_{\alpha,\beta}$ and $((\underline{R}Y)_{IF})_{\alpha,\beta}$ are $\phi$ or not $\phi$ together | $(\alpha,\beta)$ – Approximate Rough Intuitionistic Fuzzy Equivalence | $(\alpha,\beta)$ – Rough Intuitionistic Fuzzy Equivalence |

### 5.1 A general Analysis of rough equalities

In this section we provide a comparative analysis of the four types of rough equalities of intuitionistic fuzzy sets. However, the same analysis is applicable for rough equalities for fuzzy sets as well as rough equalities for crisp sets.

**5.1.1** If $((\underline{R}X)_{IF})_{\alpha,\beta} = ((\underline{R}Y)_{IF})_{\alpha,\beta}$ then either both $((\underline{R}X)_{IF})_{\alpha,\beta}$ and $((\underline{R}Y)_{IF})_{\alpha,\beta}$ are equal to $\phi$ or both are not equal to $\phi$. So, $(\alpha,\beta)$ – Rough Fuzzy Equality is a special case of $(\alpha,\beta)$ – Approximate Rough Fuzzy Equivalence and $(\alpha,\beta)$ – Approximate Rough Fuzzy Equality is a special case of $(\alpha,\beta)$ – Rough Fuzzy Equivalence. Similarly if $((\overline{R}X)_{IF})_{\alpha,\beta} = ((\overline{R}Y)_{IF})_{\alpha,\beta}$ then either $((\overline{R}X)_{IF})_{\alpha,\beta}$ and $((\overline{R}Y)_{IF})_{\alpha,\beta}$ are both equal to U or both not equal to U. So, $(\alpha,\beta)$ – Rough Fuzzy Equality is a special case of $(\alpha,\beta)$ – Approximate Rough Fuzzy Equality and $(\alpha,\beta)$ – Approximate Rough Fuzzy Equivalence is a special case of $(\alpha,\beta)$ – Rough Fuzzy Equivalence.

**5.1.2** So, it is very clear that $(\alpha,\beta)$ – Rough Fuzzy Equality is a special case of all the other three types of rough intuitionistic fuzzy approximate equalities and $(\alpha,\beta)$ – Rough Fuzzy Equivalence is a generalization of all the other three types of rough intuitionistic fuzzy approximate equalities.

**5.1.3** However, we cannot have a scale of the four types of rough intuitionistic fuzzy approximate equalities as the two types $(\alpha,\beta)$ – Approximate Rough Fuzzy Equality and $(\alpha,\beta)$ – Approximate Rough Fuzzy Equivalence have no comparison in general.

**5.1.4** The analysis in the above three cases is applicable to all the special cases, that is for rough fuzzy approximate equality and rough approximate equality.

We can represent the above analysis in the following diagram: (Here the → represents implication)

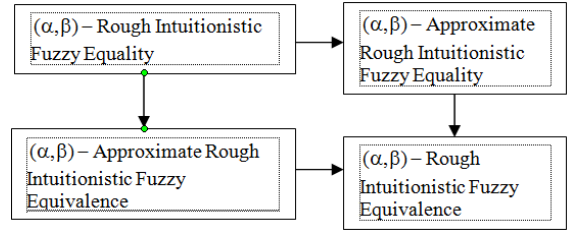

### 5.2 Examples

We illustrate here with one example, the applicability and relative efficiency of the four approximate equalities of intuitionistic fuzzy sets.

**Example 5.2.1**
Let $U = \{x_1, x_2, x_3, x_4, x_5, x_6, x_7, x_8\}$. R be an equivalence relation on U such that
$U/R = \{\{x_1, x_2\}, \{x_3, x_4, x_5, x_6\}, \{x_7, x_8\}\}$,
$X = \{(x_1, 0.2, 0.7), (x_2, 0.8, 0.1), (x_3, 0.7, 0.2), (x_4, 0.4, 0.4),$
$(x_5, 0.6, 0.3), (x_6, 0, 0.8), (x_7, 1, 0), (x_8, 0.7, 0.2)\}$ and
$Y = \{(x_1, 0, 0.8), (x_2, 0.7, 0.2), (x_3, 0.9, 0), (x_4, 0.6, 0.2),$
$(x_5, 0.4, 0.4), (x_6, 0.2, 0.7), (x_7, 0.8, 0.1), (x_8, 0.8, 0.1)\}$
be two fuzzy sets defined over U. Then we have,
$(\underline{R}X)_{IF} = \{(x_1, 0.2, 0.7), (x_2, 0.2, 0.7), (x_3, 0, 0.8), (x_4, 0, 0.8),$
$(x_5, 0, 0.8), (x_6, 0, 0.8), (x_7, 0.7, 0.2), (x_8, 0.7, 0.2)\}$.
So, taking $\alpha = 0.1, \beta = 0.8,$ we find that
$((\underline{R}X)_{IF})_{\alpha,\beta} = \{x_1, x_2\} \neq \phi.$

$(\underline{R}Y)_{IF} = \{(x_1,0,0.8),(x_2,0,0.8),(x_3,0.2,0.7),(x_4,0.2),$
$(x_5,0.2,0.7),(x_6,0.2,0.7),(x_7,0.8,0.1),(x_8,0.8,0.1)\}.$
So, $((\underline{R}Y)_{IF})_{\alpha,\beta} = \{x_3,x_4,x_5,x_6,x_7,x_8\} \neq \phi$
$X^C = \{(x_1,0.7,0.2),(x_2,0.1,0.8),(x_3,0.2,0.7),(x_4,0.4,0.4),$
$(x_5,0.3,0.6),(x_6,0.8,0),(x_7,0,1),(x_8,0.2,0.7)\}$
$(\underline{R}X^C)_{IF} = \{(x_1,0.1,0.8),(x_2,0.1,0.8),(x_3,0.2,0.7),$
$(x_4,0.2,0.7),(x_5,0.2,0.7),(x_6,0.2,0.7),(x_7,0,1),(x_8,0,1)\}$
$(\overline{R}X)_{IF} = \{(x_1,0.8,0.1),(x_2,0.8,0.1),(x_3,0.7,0.2),(x_4,0.7,0.2),$
$(x_5,0.7,0.2),(x_6,0.7,0.2),(x_7,1,0),(x_8,1,0)\}$
$((\overline{R}X)_{IF})_{\alpha,\beta} = \{x_1,x_2,x_3,x_4,x_5,x_6,x_7,x_8\} = U.$
$Y^C = \{(x_1,0.8,0),(x_2,0.2,0.7),(x_3,0,0.9),(x_4,0.2,0.6),$
$(x_5,0.4,0.4),(x_6,0.7,0.2),(x_7,0.1,0.8),(x_8,0.1,0.8)\}$
$(\underline{R}Y^C)_{IF} = \{(x_1,0.2,0.7),(x_2,0.2,0.7),(x_3,0,0.9),(x_4,0,0.9),$
$(x_5,0,0.9),(x_6,0,0.9),(x_7,0.1,0.8),(x_8,0.1,0.8)\}$
$(\overline{R}Y)_{IF} = \{(x_1,0.7,0.2),(x_2,0.7,0.2),(x_3,0.9,0),(x_4,0.9,0),$
$(x_5,0.9,0),(x_6,0.9,0),(x_7,0.8,0.1),(x_8,0.8,0.1)\}$
$((\overline{R}Y)_{IF})_{\alpha,\beta} = \{x_1,x_2,x_3,x_4,x_5,x_6,x_7,x_8\} = U.$

**Analysis 5.2.1:**
The two intuitionistic fuzzy sets are similar to each other in the sense of membership and non-membership values and we see that with the levels 0.1 and 0.8. These two intuitionistic fuzzy sets are neither rough intuitionistic fuzzy equal nor approximately rough intuitionistic fuzzy equal. But they are approximate rough intuitionistic fuzzy equivalent and rough intuitionistic fuzzy equivalent. So, the last two notions are more realistic than the other two.

**Example 5.2.2**
Let us modify example 3.1 slightly by taking
$X = \{(x_1,0,0.8),(x_2,0,0.8),(x_3,0.7,0.2),(x_4,0.4,0.4),$
$(x_5,0.6,0.3),(x_6,0.2,0.7),(x_7,1,0),(x_8,0,0.8)\}$ and
$Y = \{(x_1,0.2,0.6),(x_2,0.2,0.6),(x_3,0.9,0),(x_4,0.6,0.2),$
$(x_5,0.2,0.6),(x_6,1,0),(x_7,0,0.9),(x_8,0,0.8)\}$
Then we have,
$(\underline{R}X)_{IF} = \{(x_1,0,0.8),(x_2,0,0.8),(x_3,0.2,0.7),(x_4,0.2,0.7),$
$(x_5,0.2,0.7),(x_6,0.2,0.7),(x_7,0,0.8),(x_8,0,0.8)\}$.
So, taking $\alpha = 0.1$ and $\beta = 0.8$, we have $((\underline{R}X)_{IF})_{\alpha,\beta}$
$= \{x_3,x_4,x_5,x_6\} \neq \phi.$
$(\underline{R}Y)_{IF} = \{(x_1,0.2,0.6),(x_2,0.2,0.6),(x_3,0.2,0.6),(x_4,0.2,0.6),$
$(x_5,0.2,0.6),(x_6,0.2,0.6),(x_7,0,0.9),(x_8,0,0.9)\}.$
So, $((\underline{R}Y)_{IF})_{\alpha,\beta} = \{x_1,x_2,x_3,x_4,x_5,x_6\} \neq \phi.$

$X^C = \{(x_1,0.8,0),(x_2,0.8,0),(x_3,0.2,0.7),(x_4,0.4,0.4),$
$(x_5,0.3,0.6),(x_6,0.7,0.2),(x_7,0,1),(x_8,0.8,0)\}$
$(\underline{R}X^C)_{IF} = \{(x_1,0.8,0),(x_2,0.8,0),(x_3,0.2,0.7),(x_4,0.2,0.7),$
$(x_5,0.2,0.7),(x_6,0.2,0.7),(x_7,0,1),(x_8,0,1)\}$
$(\overline{R}X)_{IF} = \{(x_1,0,0.8),(x_2,0,0.8),(x_3,0.7,0.2),(x_4,0.7,0.2),$
$(x_5,0.7,0.2),(x_6,0.7,0.2),(x_7,1,0),(x_8,1,0)\}$
$((\overline{R}X)_{IF})_{\alpha,\beta} = \{x_3,x_4,x_5,x_6,x_7,x_8\} \neq U.$
$Y^C = \{(x_1,0.6,0.2),(x_2,0.6,0.2),(x_3,0,0.9),(x_4,0.2,0.6),$
$(x_5,0.6,0.2),(x_6,0,1),(x_7,0.9,0),(x_8,0.8,0)\}$
$(\underline{R}Y^C)_{IF} = \{(x_1,0.6,0.2),(x_2,0.6,0.2),(x_3,0,1),(x_4,0,1),$
$(x_5,0,1),(x_6,0,1),(x_7,0.8,0),(x_8,0.8,0)\}$
$(\overline{R}Y)_{IF} = \{(x_1,0.2,0.6),(x_2,0.2,0.6),(x_3,1,0),(x_4,1,0),$
$(x_5,1,0),(x_6,1,0),(x_7,0,0.8),(x_8,0,0.8)\}$
$((\overline{R}Y)_{IF})_{\alpha,\beta} = \{x_1,x_2,x_3,x_4,x_5,x_6\} \neq U.$

**Analysis 5.2.2:** We find that the two intuitionistic fuzzy sets differ much in the membership values of their elements, particularly for the elements $x_6$ and $x_7$. Still the two sets are rough intuitionistic fuzzy equivalent as per definition. This is the flexibility provided by this characterization in the upper approximation. But, as $(\overline{R}X)_{\alpha,\beta} \neq (\overline{R}Y)_{\alpha,\beta}$, the two fuzzy sets are not approximately rough intuitionistic fuzzy equivalent. So, once again we conclude that approximate rough intuitionistic fuzzy equivalence is more realistic than the other three types of intuitionistic fuzzy approximate equalities.

5.3 Properties of approximate equalities of rough intuitionistic fuzzy sets

Properties similar to the basic properties of rough sets hold true for rough intuitionistic fuzzy sets. The following four properties are to be used by us in establishing the properties of approximate equalities of intuitionistic fuzzy sets.
Let R be an equivalence relation defined over U and $X, Y \in IF(U)$. Then for any $(\alpha, \beta) \in J$, we have

(5.3.1) $((\underline{R}(X \cap Y))_{IF})_{\alpha,\beta} = ((\underline{R}(X))_{IF} \cap (\underline{R}(Y))_{IF})_{\alpha,\beta}$

(5.3.2) $((\underline{R}(X \cup Y))_{IF})_{\alpha,\beta} \supseteq ((\underline{R}(X))_{IF} \cup (\underline{R}(Y))_{IF})_{\alpha,\beta}$

(5.3.3) $((\overline{R}(X \cap Y))_{IF})_{\alpha,\beta} \subseteq ((\overline{R}(X))_{IF} \cap (\overline{R}(Y))_{IF})_{\alpha,\beta}$

(5.3.4) $((\overline{R}(X \cup Y))_{IF})_{\alpha,\beta} = ((\overline{R}(X))_{IF} \cup (\overline{R}(Y))_{IF})_{\alpha,\beta}$

We provide the proofs of (5.3.1) and (5.3.2) only. The other two proofs are similar.

**Proof of (5.3.1)**
For any $u \in U$, we have

$$M(\underline{R}(X \cap Y))_{IF}(u) = \inf_{y \in [u]_R} M(X \cap Y)(y)$$

$$= \inf_{y \in [u]_R} \min\{MX(y), MY(y)\}$$

$$= \min\{\inf_{y \in [u]_R} MX(y), \inf_{y \in [u]_R} MY(y)\}$$

$$= \min\{M\underline{R}X(u), M\underline{R}Y(u)\} = M((\underline{R}X \cap \underline{R}Y))_{IF}(u).$$

So, $(M(\underline{R}(X \cap Y))_{IF}(u))_{\alpha,\beta} = ((M(\underline{R}X \cap \underline{R}Y))_{IF}(u))_{\alpha,\beta}$

Also,

$$N(\underline{R}(X \cap Y))_{IF}(u) = \sup_{y \in [u]_R} N(X \cap Y)(y)$$

$$= \sup_{y \in [u]_R} \max\{NX(y), NY(y)\}$$

$$= \max\{\sup_{y \in [u]_R} NX(y), \sup_{y \in [u]_R} NY(y)\}$$

$$= \max\{N(\underline{R}X)_{IF}(u), N(\underline{R}Y)_{IF}(u)\} = N(\underline{R}X \cap \underline{R}Y)_{IF}(u).$$

So, $(N(\underline{R}(X \cap Y))_{IF}(u))_{\alpha,\beta} = (N(\underline{R}X \cap \underline{R}Y)_{IF}(u))_{\alpha,\beta}$

Hence $(5.3.1)$ holds true.

**Proof of (5.3.2)**
Since $X \subseteq X \cup Y$ and $Y \subseteq X \cup Y$ it follows that

$(\underline{R}X)_{IF} \subseteq (\underline{R}(X \cup Y))_{IF}$ and $(\underline{R}Y)_{IF} \subseteq (\underline{R}(X \cup Y))_{IF}$.

Hence, $(\underline{R}X)_{IF} \cup (\underline{R}Y)_{IF} \subseteq (\underline{R}(X \cup Y))_{IF}$.

So, $((\underline{R}X)_{IF} \cup (\underline{R}Y)_{IF})_{\alpha,\beta} \subseteq ((\underline{R}(X \cup Y))_{IF})_{\alpha,\beta}$.

To show that the inclusion can actually be strict we provide one simple example. Here, we are taking $\beta = 1 - \alpha$ and $\alpha = 0$, which is enough to establish our claim. Let $U = \{x_1, x_2, x_3, x_4, x_5\}$ and R be such that $U/R = \{\{x_1, x_2, x_4\}, \{x_3, x_5\}\}$.

$X = \{(x_1, 0.1, 0.8), (x_2, 0.5, 0.4), (x_3, 0.4, 0.4),$
$(x_4, 0.2, 0.7), (x_5, 0.6, 0.3)\}$ and
$Y = \{(x_1, 0.3, 0.6), (x_2, 0.2, 0.6), (x_3, 0.6, 0.3),$
$(x_4, 0.7, 0.2), (x_5, 0.9, 0.1)\}$.

$M(\underline{R}(X \cup Y))_{IF}(x_4) = M(\underline{R}(X \cup Y))_{IF}(x_2)$

$= M(\underline{R}(X \cup Y))_{IF}(x_1) = \inf_{y \in [x_1]_R} \text{Max}\{MX(y), MY(y)\}$

$= \inf\{\max(0.1, 0.3), \max(0.5, 0.2), \max(0.2, 0.7)\}$

$= \inf\{0.3, 0.5, 0.7\} = 0.3$.

Similarly, $M(\underline{R}(X \cup Y))_{IF}(x_3) = M(\underline{R}(X \cup Y))_{IF}(x_5) = 0.6$.

$N(\underline{R}(X \cup Y))_{IF}(x_4) = N(\underline{R}(X \cup Y))_{IF}(x_2)$

$= N(\underline{R}(X \cup Y))_{IF}(x_1) = \sup_{y \in [x_1]_R} \text{Min}\{NX(y), NY(y)\}$

$= \sup\{\min(0.8, 0.6), \min(0.4, 0.6), \min(0.7, 0.2)\}$

$= \sup\{0.6, 0.4, 0.2\} = 0.6$.

Similarly, $N(\underline{R}(X \cup Y))_{IF}(x_3) = N(\underline{R}(X \cup Y))_{IF}(x_5) = 0.3$.

So, $(\underline{R}(X \cup Y))_{IF} = \{(x_1, 0.3, 0.6), (x_2, 0.3, 0.6), (x_3, 0.6, 0.3),$
$(x_4, 0.3, 0.6), (x_5, 0.6, 0.3)\}$.

### 5.4 General properties

The definitions of bottom R-inclusion, top R-inclusion and R-inclusion for crisp sets and fuzzy sets can be extended in a natural way to intuitionistic fuzzy sets as follows.

**Definition 5.4.1:** Let R be an equivalence relation defined on U and X, Y $\in$ IF(U). Then
(i) We say that X is bottom (R, $\alpha,\beta$)-included in Y $\left(X \subseteq_{(R,\alpha,\beta)} Y\right)$ if and only if $((\underline{R}X)_{IF})_{\alpha,\beta} \subseteq ((\underline{R}Y)_{IF})_{\alpha,\beta}$

(ii) We say that X is top (R, $\alpha,\beta$)-included in Y $\left(X \tilde{\subset}_{(R,\alpha,\beta)} Y\right)$ if and only if $((\overline{R}X)_{IF})_{\alpha,\beta} \subseteq ((\overline{R}Y)_{IF})_{\alpha,\beta}$.

(iii) We say that X is (R, $\alpha,\beta$)-included in Y $\left(X \tilde{\subseteq}_{(R,\alpha,\beta)} Y\right)$ if and only if $X \subseteq_{R,\alpha,\beta} Y$ and $X \tilde{\subset}_{R,\alpha,\beta} Y$

**Definition 5.4.2:** (i) we say X, Y $\in$ IF(U) are $(\alpha,\beta)$− bottom rough comparable if and only if $X \subseteq_{R,\alpha,\beta} Y$ or $Y \subseteq_{R,\alpha,\beta} X$ holds.

(ii) We say X, Y $\in$ IF(U) are $(\alpha,\beta)$− top rough comparable if and only if $X \tilde{\subset}_{R,\alpha,\beta} Y$ or $Y \tilde{\subset}_{R,\alpha,\beta} X$ holds.

(iii) We say X, Y $\in$ IF(U) are $(\alpha,\beta)$−rough comparable if and only if X and Y are both top and bottom rough $(\alpha,\beta)$−comparable.

**Definition 5.4.3:** We say that two intuitionistic fuzzy sets X and Y are $(\alpha,\beta)$− bottom (rough equal/approximately rough equal, rough equivalent/approximately rough equivalent) if and only if

$(((\underline{R}X)_{IF})_{\alpha,\beta} = ((\underline{R}Y)_{IF})_{\alpha,\beta}, ((\underline{R}X)_{IF})_{\alpha,\beta}$ and $((\underline{R}Y)_{IF})_{\alpha,\beta}$

are equal to φ or not equal to φ together ).

**Definition 5.4.4:** We say that two intuitionistic fuzzy sets X and Y are $(\alpha,\beta)$ – top (rough equal/approximately rough equal, rough equivalent/approximately rough equivalent) iff $((\overline{R}X)_{IF})_{\alpha,\beta} = ((\overline{R}Y)_{IF})_{\alpha,\beta}$, $((\underline{R}X)_{IF})_{\alpha,\beta}$ and $((\underline{R}Y)_{IF})_{\alpha,\beta}$ are equal to U or not equal to U together ).

We state below the properties of some of the four types of approximate $(\alpha,\beta)$ – rough fuzzy equalities below with out proof. The proofs are similar to the approximate rough equality cases [13].

In the properties below, we use A is $(\alpha,\beta)$-bottom related to B to mean either $((\underline{R}A)_{IF})_{\alpha,\beta} = ((\underline{R}B)_{IF})_{\alpha,\beta}$ or $((\underline{R}A)_{IF})_{\alpha,\beta}$ and $((\underline{R}B)_{IF})_{\alpha,\beta}$ are φ or not φ together.

Also, we use

A is $(\alpha,\beta)$-top related to B to mean that either $((\overline{R}A)_{IF})_{\alpha,\beta} = ((\overline{R}B)_{IF})_{\alpha,\beta}$

or $((\overline{R}A)_{IF})_{\alpha,\beta}$ and $((\overline{R}B)_{IF})_{\alpha,\beta}$ are U or not U together.

**Property 5.4.1:** (i) If $X \cap Y$ is $(\alpha,\beta)$-bottom related to both X and Y then X is $(\alpha,\beta)$-bottom related to Y.

(ii) The converse of (i) is not true in general and an additional condition that is sufficient but not necessary for the converse to be true is that X and Y are $(\alpha,\beta)$ – bottom rough comparable.

**Proof:** There are two cases.

**Case-I**

If $(\underline{R}(X \cap Y)_{IF})_{\alpha,\beta} = ((\underline{R}X)_{IF})_{\alpha,\beta}$ and $(\underline{R}(X \cap Y)_{IF})_{\alpha,\beta}$

$= ((\underline{R}Y)_{IF})_{\alpha,\beta}$ then $((\underline{R}X)_{IF})_{\alpha,\beta} = ((\underline{R}Y)_{IF})_{\alpha,\beta}$.

Then the proof follows.

**Case-II**

If $(\underline{R}(X \cap Y)_{IF})_{\alpha,\beta}$ and $((\underline{R}X)_{IF})_{\alpha,\beta}$ are φ or (≠ φ) together

then and $(\underline{R}(X \cap Y)_{IF})_{\alpha,\beta}$ and $((\underline{R}Y)_{IF})_{\alpha,\beta}$ are φ or (≠ φ)

together respectively. So that $((\underline{R}X)_{IF})_{\alpha,\beta}$ and $((\underline{R}Y)_{IF})_{\alpha,\beta}$

are φ or (≠ φ) together. Hence the proof follows.

Conversely,

If $((\underline{R}X)_{IF})_{\alpha,\beta} = ((\underline{R}Y)_{IF})_{\alpha,\beta}$ then by (5.3.1)

$((\underline{R}(X \cap Y))_{IF})_{\alpha,\beta} = ((\underline{R}X)_{IF})_{\alpha,\beta} \cap ((\underline{R}Y)_{IF})_{\alpha,\beta}$

$= ((\underline{R}X)_{IF})_{\alpha,\beta} = ((\underline{R}Y)_{IF})_{\alpha,\beta}$.

So, the proof follows.

Next, if

$(\underline{R}X)_{\alpha,\beta}$ and $(\underline{R}Y)_{\alpha,\beta}$ are both φ then $(\underline{R}(X \cap Y))_{\alpha,\beta}$

$= (\underline{R}X)_{\alpha,\beta} \cap (\underline{R}Y)_{\alpha,\beta} = \phi$. So, the converse is true.

However, if both $(\underline{R}X)_{\alpha,\beta}$ and $(\underline{R}Y)_{\alpha,\beta}$ are not equal

to φ then $(\underline{R}(X \cap Y))_{\alpha,\beta}$ may be φ. So that the result may

not be true. In addition, if $(\underline{R}X)_{\alpha,\beta}$ and $(\underline{R}Y)_{\alpha,\beta}$ are

bottom $(\alpha,\beta)$-comparable then $(\underline{R}(X \cap Y))_{\alpha,\beta} = (\underline{R}X)_{\alpha,\beta}$ or $(\underline{R}Y)_{\alpha,\beta}$ as the case may be. So, it is also not φ and the proof follows.

**Property 5.4.2:** (i) If $X \cup Y$ is $(\alpha,\beta)$-top related to both X and Y then X is $(\alpha,\beta)$-top related to Y.

(ii) The converse of (i) is not true in general and an additional condition that is sufficient but not necessary for the converse to be true is that X and Y are top $(\alpha,\beta)$ – rough comparable.

**Proof:** Similar to that of property 5.4.1.

**Property 5.4.3:** (i) If X is $(\alpha,\beta)$-top related to X' and Y is $(\alpha,\beta)$-top related to Y' then it may or may not be true that $X \cup Y$ is $(\alpha,\beta)$-top related $X' \cup Y'$.

(ii) A sufficient but not necessary condition for the result in (i) to be true is that X and Y are $(\alpha,\beta)$ – top rough comparable and X' and Y' are $(\alpha,\beta)$ – top rough comparable.

**Proof: Proof of (i)** There are two cases

**Case-I**

Suppose $((\overline{R}X)_{IF})_{\alpha,\beta} = ((\overline{R}X')_{IF})_{\alpha,\beta}$ and

$((\overline{R}Y)_{IF})_{\alpha,\beta} = ((\overline{R}Y')_{IF})_{\alpha,\beta}$. Then using (5.3.4),

it follows that $((\overline{R}(X \cup Y))_{IF})_{\alpha,\beta} = ((\overline{R}(X' \cup Y'))_{IF})_{\alpha,\beta}$.

**Case-II**

Suppose both $((\overline{R}X)_{IF})_{\alpha,\beta}, ((\overline{R}X')_{IF})_{\alpha,\beta}$ and $((\overline{R}Y)_{IF})_{\alpha,\beta}$,

$((\overline{R}Y')_{IF})_{\alpha,\beta}$ are U or not U together then

it may happen that one of $((\overline{R}(X \cup Y))_{IF})_{\alpha,\beta}$

and $((\overline{R}(X' \cup Y'))_{IF})_{\alpha,\beta}$ is U and the other one is not U.

So that the equality does not hold.

**Proof of (ii)**

But, if X and Y are $(\alpha,\beta)$-top rough comparable and

X' and Y' are also $(\alpha,\beta)$-top rough comparable then

$((\overline{R}(X \cup Y))_{IF})_{\alpha,\beta}$ is equal to one of $((\overline{R}(X))_{IF})_{\alpha,\beta}$ and

$(\overline{R}(Y))_{\alpha,\beta}$ and so is the case for $((\overline{R}(X' \cup Y'))_{IF})_{\alpha,\beta}$.

So, the equality follows.

The claim that the condition is not necessary follows from the fact that it is not necessary in the base case [13].

The proofs of the following properties can be carried out similarly. We only state them.

**Property 5.4.4:** (i) X is $(\alpha,\beta)$-bottom related X' and Y is $(\alpha,\beta)$-bottom related to Y' may or may not imply that $X \cap Y$ is $(\alpha,\beta)$-bottom related to $X' \cap Y'$

(ii) A sufficient but not necessary condition for the result in (i) to be true is that X and Y are $(\alpha,\beta)$–bottom rough comparable and $X'$ and $Y'$ are $(\alpha,\beta)$–bottom rough comparable.

**Property 5.4.5:** (i) X is $(\alpha,\beta)$-top related to Y may or may not imply that $X \cup (-Y)$ is $(\alpha,\beta)$-top related to U.

(ii) A sufficient but not necessary condition for result in (i) to hold is that X and Y are $(\alpha,\beta)$-bottom rough equal.

**Property 5.4.6:** (i) X is $(\alpha,\beta)$-bottom related to Y may not imply that $X \cap (-Y)$ is $(\alpha,\beta)$-bottom related to $\phi$.

(ii) A sufficient but not necessary condition for the result in (i) to hold true is that X and Y are $(\alpha,\beta)$-top rough equal.

**Property 5.4.7:** If $X \subseteq Y$ and Y is $(\alpha,\beta)$-bottom related to $\phi$ then X $(\alpha,\beta)$-bottom related to $\phi$.

**Property 5.4.8:** If $X \subseteq Y$ and X is $(\alpha,\beta)$-top related to U then Y is $(\alpha,\beta)$-top related to U.

**Property 5.4.9:** X is $(\alpha,\beta)$-top related to Y if and only if –X is $(\alpha,\beta)$-bottom related to –Y.

**Property 5.4.10:** If X is $(\alpha,\beta)$-bottom related to $\phi$ and Y is $(\alpha,\beta)$-bottom related to $\phi$ then $X \cap Y$ is $(\alpha,\beta)$-bottom related to $\phi$.

**Property 5.4.11:** If X is $(\alpha,\beta)$-top related to U or Y is $(\alpha,\beta)$-top related to U then $X \cup Y$ is $(\alpha,\beta)$-top related to U.

## 5.5 Replacement properties for rough intuitionistic fuzzy equivalence

We state below the properties which are obtained by interchanging bottom rough intuitionistic fuzzy equalities with top rough intuitionistic fuzzy equalities. The proofs are similar to the properties in section 5.4. So, we omit them.

**Property 5.5.1:** (i) If $X \cap Y$ is $(\alpha,\beta)$-top related to both $X$ and $Y$ then X is $(\alpha,\beta)$-top related to $Y$.

(ii) The converse of (i) is not true in general and an additional condition that is sufficient but not necessary for the converse to be true is that equality holds in (3.2.3).

**Property 5.5.2:** (i) $X \cup Y$ is $(\alpha,\beta)$-bottom related to $X$ and $X \cup Y$ is $(\alpha,\beta)$-bottom related to $Y$ then $X$ is $(\alpha,\beta)$-bottom related to $Y$

(ii) The converse of (i) is not true in general and an additional condition that is sufficient but not necessary for the converse to be true is that equality holds in (3.2.2).

**Property 5.5.3:** (i) $X$ is $(\alpha,\beta)$-bottom related to $X'$ and $Y$ is $(\alpha,\beta)$-bottom related to $Y'$ may not imply $X \cup Y$ is $(\alpha,\beta)$-bottom related to $X' \cup Y'$.

(ii) A sufficient but not necessary condition for the conclusion of (i) to hold is that equality holds in (3.2.3).

**Property 5.5.4:** (i) $X$ is $(\alpha,\beta)$-top related to $X'$ and $Y$ is $(\alpha,\beta)$-top related to $Y'$ may not necessarily imply that $X \cap Y$ is $(\alpha,\beta)$-top related to $X' \cap Y'$.

(ii) A sufficient but not necessary condition for the conclusion in (i) to hold is that equality holds in (3.2.2).

**Property 5.5.5:** X is $(\alpha,\beta)$-bottom related to $Y$ may or may not imply $X \cup -Y$ is $(\alpha,\beta)$-bottom related to $U$.

**Property 5.5.6:** $X$ is $(\alpha,\beta)$-top related to $Y$ may or may not imply $X \cap -Y$ is $(\alpha,\beta)$-top related to $\phi$.

Properties (5.4.7) to (5.4.11) hold true under replacements.

## 5.6 Real life Examples

**Example 1:** Let us consider the universal set of all faculties in colleges in a state. We define xRy if and only if x and y belong to the same college. This is an equivalence relation and decomposes the set of faculties into equivalence classes, which are colleges in the state.

We define two intuitionistic fuzzy sets G and Y over the universe as comprising of "good" faculty and "young" faculty. These two concepts can be defined through intuitionistic fuzzy membership values. For example a faculty x can be considered as a member of G as (x, 0.6, 0.2) and the same faculty can be defined as a member of Y as (x, 0.7, 0.1).

**Case I.** Suppose $(\underline{R}G)_{IF} = (\underline{R}Y)_{IF}$. It implies that the colleges for which all the faculties are good also have all the faculties young.

**Case II.** Suppose $(\underline{R}G)_{IF}$ and $(\underline{R}Y)_{IF}$ are $\phi$ or not $\phi$ together. This implies that either there is no college which contain all good faculty or young faculty or there are some colleges which contain all good faculties or all young faculties. Unlike case-I, here the set of colleges may not be same.

**Case III.** Suppose $(\overline{R}G)_{IF} = (\overline{R}Y)_{IF}$. It implies that the set of colleges which have at least one good faculty is same as the set of colleges which have at least one young faculty.

**Case IV.** Suppose $(\overline{R}G)_{IF}$ and $(\overline{R}Y)_{IF}$ are U or not U together. If both are U then it implies that all the colleges have at least one good faculty as well as at least one young faculty. If both are not U it implies there are some colleges which do not have any young faculty and there some colleges which do not have any good faculty. Unlike case-III, here the same colleges may not have this feature.

**Note 5.6.1:** We can have combinations of the above four cases to get the four types of approximate equalities provided in table 5.1. For example, if we have both case I and case III are true then we can conclude that the colleges for which all the faculties are good also have all the faculties young and those colleges have some faculties young also have some faculties who are good.

**Note 5.6.2:** It may be noted that we can add control values $\alpha$ and $\beta$ to discuss the above cases and make them refined. For example, if we want a goodness factor of 0,5 and also a youth factor of also 0.5 we can do so by taking $\alpha = 0.5$ and also we may like to take $\beta = 0.1$. This will enable us to consider only those faculties who are having high goodness factor as well as high youth factor for comparison. Also, we can reduce the factor of not being good or not being young to filter out faculties having high negative values.

## 6. Conclusions

In this paper we introduced the concept of leveled approximate equality by taking rough intuitionistic fuzzy sets instead of rough fuzzy sets considered in [15] in defining approximate equality. This notion has two advantages over the previous model. First, it considers the approximate equality of intuitionistic fuzzy sets instead of fuzzy sets. Second, we continue with the notion of bi-leveled approximate equality, which provides a pair of threshold values for the approximate equality. This provides a double control with the user to specify the desired levels of equality needed. So that, according to the requirement these values can be adjusted depending upon the situation. These notions are more realistic than those considered earlier by Novotny and Pawlak [8, 9, 10] and Tripathy et al [11, 12, 13, 15]. Also, we provided two real life examples to demonstrate the use of rough fuzzy equalities. Some properties of the four types of rough intuitionistic fuzzy equalities are established. We provided some theoretical examples to compare the efficiencies of the rough intuitionistic fuzzy equalities.

**First Author** Dr.B.K.Tripathy is a senior professor in the school SCSE of VIT University, Vellore, India since 2007. He has produced 12 PhDs so far and has published over 140 technical papers in various international/National Journals and International/National Conference proceedings/Springer book chapters. He is in the editorial board of several journals and is a reviewer of many other journals including some Elsevier, Springer and World Scientific publications. He is a member/life member of international bodies including IEEE,ACEEE, IST, ACM, IRSS, AISTC, WSEAS, CSI and IMS. His current fields of research interests include Fuzzy set theory and systems, Rough sets and knowledge engineering, Granular computing, Social Networks, Data Clustering Techniques, List theory and Concept Analysis.

**Second Author** G. K. Panda has recognized contributions in Rough set, Knowledge Engineering and Social Networking. He has published, reviewed several research articles in refereed intl. conferences and journals and a member of ISAI, OITS, ISTE etc. At present he is an Associate Professor in M.I.T.S., India.